\documentclass[11pt,letterpaper]{article}
\usepackage{emnlp2016}
\usepackage{times}
\usepackage{latexsym}
\usepackage{times}
\usepackage{url}
\usepackage{latexsym}
\usepackage{epsfig}
\usepackage{array}
\usepackage{color}
\usepackage{multirow}
\usepackage{url}
\usepackage{subfig}
\usepackage{algorithmic}
\usepackage{algorithm}
\usepackage{multirow}
\usepackage{rotating}
\usepackage{booktabs}
\usepackage{makecell}
\usepackage{comment}
\usepackage{amsmath,amsfonts,amssymb}
\usepackage{tikz}
\usepackage{verbatim}
\usepackage{caption}
\usepackage{booktabs}
\usepackage{ctable}
\usepackage{makecell}
\usepackage{array} 
\usepackage{varwidth} 
\usepackage{stmaryrd}
\usepackage{dsfont}
\usepackage{todonotes}

\emnlpfinalcopy

\title{Continuously Learning Neural Dialogue Management}

\author{
Pei-Hao Su, Milica Ga{\v s}i{\' c}, Nikola Mrk{\v s}i{\' c}, Lina Rojas-Barahona, \\ {\bf Stefan Ultes, David Vandyke, Tsung-Hsien Wen and Steve Young} \\
Department of Engineering,
University of Cambridge,
Cambridge, UK\\
\texttt{\{phs26, mg436, nm480, lmr46, su259, djv27, thw28, sjy\}@cam.ac.uk} \\
}
\date{}

\begin{document}

\maketitle

\begin{abstract}
We describe a two-step approach for dialogue management in task-oriented spoken dialogue systems. A unified neural network framework is proposed to enable the system to first learn by supervision from a set of dialogue data and then continuously improve its behaviour via reinforcement learning, all using gradient-based algorithms on one single model. The experiments demonstrate the supervised model's effectiveness in the corpus-based evaluation, with user simulation, and with paid human subjects. The use of reinforcement learning further improves the model's performance in both interactive settings, especially under higher-noise conditions.

\end{abstract}

\section{Introduction} \label{sec:intro}

Developing a robust Spoken Dialogue System (SDS) traditionally requires a substantial amount of hand-crafted rules combined with various statistical components. 
In a task-oriented SDS, teaching a system how to respond appropriately is non-trivial. 
More recently, this {\it dialogue management} task has been formulated as a reinforcement learning (RL) problem which can be automatically optimised through human interaction \cite{levin1997stochastic,roy2000spoken,POMDP_williams,jurvcivcek2011natural,POMDP-review}. In this framework, the system learns by a {\it trial and error} process governed by a potentially delayed learning objective, a {\it reward function}, that determines dialogue success \cite{el2014task,Su_2015,vandyke15a,su:2016:acl}. 
To enable the system to be trained on-line, sample-efficient learning algorithms have been proposed \cite{GPRL,KTD} which can learn policies from a minimal number of dialogues.
However, even with such methods, performance is still poor in the early training stages, and this can impact negatively on the user experience. For these and other reasons, most commercial systems still hand-craft the dialogue management to ensure its stability.

Supervised learning (SL) has also been used in dialogue research
where a policy is trained to produce an example response given the dialogue state. Wizard-of-Oz (WoZ) methods \cite{Kelley84,dahlback1993wizard} have been widely used for collecting domain-specific training corpora.
Recently an emerging line of research has focused on training a network-based dialogue model, mostly in text-input schemes \cite{vinyals2015neural,serban2015hierarchical,wenN2N16,fb_n2n}.
These systems were directly trained on past dialogues without detailed specification of the internal dialogue state.
However, there are two key limitations of using the SL approach for SDS. Firstly, the effects of selecting an action on the future course of the dialogue are not considered. Secondly, there may be a very large number of dialogue states for which an appropriate response must be generated. Hence, the SL training set may lack sufficient coverage. Another issue is that there is no reason to suppose a human wizard is acting optimally, especially at high noise levels.
These problems exacerbate in larger domains where multi-step planning is needed. Thus, learning to mimic a human wizard does not necessary lead to optimal behaviour.

To get the best of both SL- and RL-based dialogue management, this paper describes a network-based model which is initially trained with a supervised spoken dialogue dataset. Since the training data may be mismatched to the deployment environment, the model is further improved by RL in interaction with a simulated user or human users. The advantage of the proposed framework is that a single model can be trained using both SL and RL without modifying the system architecture. This resembles the training process used in AlphaGo \cite{silver2016mastering} for the game of Go.
In addition, unlike most of the-state-of-the-art RL-based dialogue systems \cite{GPRL,cuayahuitl2015strategic} which operate on a constrained set of {\it summary} actions to limit the policy space and minimise expensive training costs, our model operates on a full action set.

\section{Neural Dialogue Management} \label{sec:models}
\label{sec:architecture}
\begin{figure}[t]
\centerline{\includegraphics[scale=0.25]{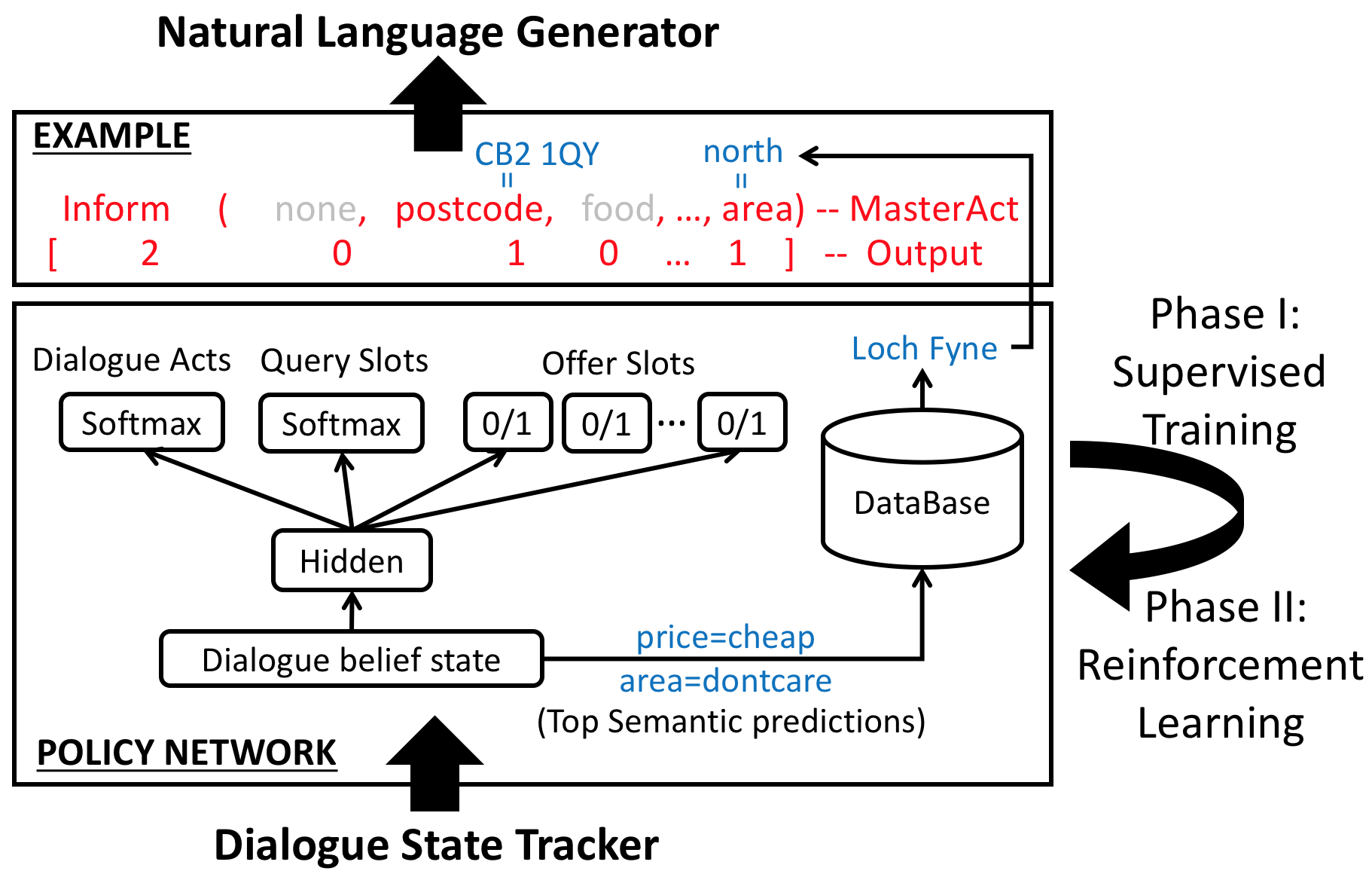}}
\caption{Network-based System Architecture.} 
\label{fig:architecture}
\end{figure}

The proposed framework addresses the dialogue management component in a modular SDS. As depicted in Figure \ref{fig:architecture}, the input to the model is the belief state $s$ which encodes the understood user intents along with the dialogue history \cite{Henderson2014b,Mrksic:15}, and the output is the master dialogue action $a$ that decides the semantic reply. This is subsequently passed to the natural language generator \cite{wensclstm15}.

Dialogue management is represented as a {\bf Policy Network}, a neural network with one hidden layer exploiting {\it tanh} non-linearities, an output layer consisting of two softmax partitions and six sigmoid partitions. For the softmax outputs, one is for predicting \textit{\textbf{DiaAct}}, a multi-class label over five dialogue acts: \{\texttt{request}, \texttt{offer}, \texttt{confirm}, \texttt{select}, \texttt{bye}\}, and the other for predicting \textit{\textbf{Query}}, containing four options for the search constraint: \{\texttt{food}, \texttt{pricerange}, \texttt{area}, \texttt{none}\}.
\textit{\textbf{Query}} options only matter if the dialogue act in \{\texttt{request}, \texttt{confirm}, \texttt{select}\} is used.
The sigmoid partitions are for \textit{\textbf{Offer}}, each of which is used to determine a binary prediction when making system offer\footnote{System-offer slots are slots the system can mention, such as area, phone number and postcode.}.

Given the system's understanding of the user, the model's role is to determine {\it what} the intent of the system response should be and which {\it slot} to talk about. The exact {\it value} in each slot is decided by a separate database parser, where the query is the top prediction of each user-informable slot\footnote{User-informable slots are slots used by the user to constrain the search, such as area and price range.} from the dialogue state tracker and the output is a matched entity. This output forms the system's semantic reply, the {\it master dialogue action}. 

\subsection{Phase I: Supervised Learning}
\label{sec:sl}
In the first phase, the policy network is trained on corpus data. This data may come from WoZ collection or from interactions between users and an already existing policy, such as a hand-crafted baseline. The objective here is to `mimic' the response behaviour within the supervised data. 

The training objective for each sample is to minimise a joint cross-entropy loss $\mathcal{L}(\theta)$ between model action labels $y$ defined in \S \ref{sec:architecture} and predictions $p$:
\begin{equation}
\label{eq:loss}
\begin{split}
\mathcal{L}(\theta) = \sum\limits_{k \in \{d_a, q, O_s\}} H(y_k, p_k),
\end{split}
\end{equation}
where DiaAct $d_a$ and Query $q$ outputs are categorical distributions, and the Offer set $O_s$ 
contains six binary offer slots.
$\theta$ are the network parameters.

\subsection{Phase II: Reinforcement Learning}
\label{sec:rl}

The policy trained in phase I on a fixed dataset may not generalise well. In spoken dialogue, the noise level may vary across conditions and thus can significantly affect performance.
Hence, the second phase of the training pipeline aims at improving the SL trained policy network by further training using policy-gradient based RL. The model is given the freedom to select any combination of master action.
The training objective is to find a parametrised policy $\pi_\theta$ that maximises the expected reward $J(\theta)$ of a dialogue with $T$ turns:
$J(\theta) = E \left[ \sum_{t=1}^{T} \gamma^t r(s_t, a_t) \middle|  \pi_{\theta} \right]$,
where $\gamma$ is the discount factor
and $r(s_t, a_t)$ is the reward when taking master action $a_t$ in dialogue state $s_t$. Note that the structure and initial weights of $\pi_\theta$ are fixed by the SL pre-training phase, since the RL training aims to improve the SL trained model.

Here a batch algorithm is adopted, and all transitions are sampled under the current policy. 
At each update iteration, $N$ episodes were generated, where the $i$th episode consists of a set of transition tuples $\{ (s_t^i, a_t^i, r_t^i)\}_{t=0}^{T_i}$.
The estimated gradient is estimated using the likelihood ratio trick: 
\begin{equation}
{\nabla_\theta J(\theta)} = \frac{1}{N{T_i}} \sum_{i=1}^N \sum_{t=0}^{T_i} \nabla_\theta \log \pi(a_t^i|s_t^i; \theta) R_t^i,
\end{equation}
where $R_t^i = \sum_{t'=t}^{T_i} \gamma^{t'-t} r_{t'}^i$ is the cumulative return from time-step $t$ to $T_i$.
Gradient descent is, however, slow and has poor convergence properties.

{\it Natural gradient} \cite{amari1998natural} improves upon the above 'vanilla' gradient method by computing an ascent direction that
approximately ensures a small change in the policy distribution.
This direction is $ w = F(\theta)^{-1} {\nabla_\theta J(\theta)}$, where $F(\theta)$ is the Fisher information matrix (FIM). Based on this, \newcite{peters2006policy} developed the Natural Actor-Critic (NAC). In its episodic case (eNAC), the FIM does not need to be explicitly computed to obtain the natural gradient $w$. eNAC uses a least square method:
\begin{equation}
R_n = \left[ \sum_{t=0}^T \nabla_\theta \log \pi(a_t^i|s_t^i; \theta)^T \right] \cdot w + C, 
\end{equation}
where $C$ is a constant and $\forall n \in \{1,...,N\}$ an analytical solution can be obtained. For larger models with more parameters, a truncated variant \cite{schulman2015trust} can also be used to practically calculate the natural gradient.

{\it Experience replay} \cite{lin1992self} is utilised to randomly sample mini-batches of experiences from a reply pool $\mathcal{P}$. This increases data efficiency by re-using experience samples in multiple updates and reduces the data correlation.
As the gradient is highly correlated with the return $R$, to ensure stable training, a unity-based {\it reward normalisation} is adopted to normalise the total return $R_n$ between 0 and 1.

\section{Experimental Results} \label{sec:exp}

The target application is a live telephone-based SDS providing restaurant information for the Cambridge (UK) area. The domain consists of approximately 150 venues, each having 6 slots out of which 3 can be used by the system to constrain the search (food-type, area and price-range) and 3 are informable properties (phone-number, address and postcode) available once a database entity has been found. 

The model was implemented using the Theano library \cite{2016arXiv160502688short}. The size of the hidden layer was set to 32 and all the weights were randomly initialised between -0.1 and 0.1.

\subsection{Supervised Learning on Corpus Data} \label{sec:supervised}

A corpus consisting of 720 user dialogues in the Cambridge restaurant domain was split into 4:1:1 for training, validation and testing. This corpus was collected via the Amazon Mechanical Turk (AMT) service, where paid subjects interacted through speech with a well-behaved dialogue system as proposed in \cite{su:2016:acl}.
The raw data contains the top N speech recognition (ASR) results which were passed to a rule-based semantic decoder and the {\it focus} belief state tracker \cite{Henderson2014a} to obtain the belief state that serves as the input feature to the proposed policy network. 
The turn-level labels were tagged according to \S \ref{sec:architecture}.
Adagrad \cite{duchi2011adaptive} per dialogue was used during backpropagation to train each model based on the objective in Equation \ref{eq:loss}. 
To prevent over-fitting, early stopping was applied based on the held-out validation set.

Table \ref{tab:sl} shows the weighted F-1 scores computed on the test set for each label. We can clearly see that the model accurately determines the type of reply (DiaAct) and generally provides the right information (Offer). The hypothesised reason for the lower accuracy of \textit{\textbf{Query}} is that the SL training data contains robust ASR results and thus the system examples contain more offers and less queries. This can be mitigated with a larger dataset covering more diverse situations, or improved via an RL approach.

\begin{table}[h]
  \caption{Model performance based on F-measure.}
  \label{tab:sl}
  \centering
  	\begin{tabular}{cccc}
    	\toprule
    	{\bf Output}	&	DiaAct	&	Query	&	Offer \\
   		\midrule
   		{\bf F-1}	&	97.73	&	87.39	&	92.51 \\
   		\bottomrule
 	\end{tabular}
  \vspace{-3mm}
\end{table}

\subsection{Policy Network in Simulation} \label{sec:simulate}

\begin{figure}[t]
\centerline{\includegraphics[scale=0.45]{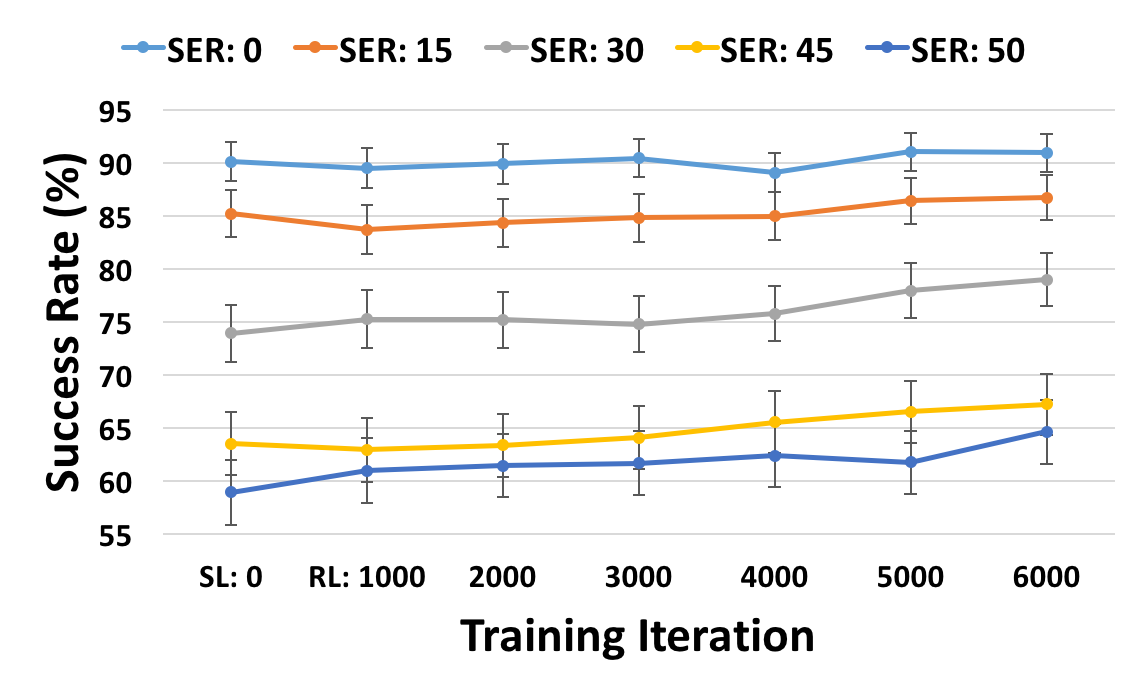}}
\caption{The success rate of the policy network in user simulation under various semantic error rates trained with SL and further improved via RL.} 
\label{fig:SLvsRL}
\end{figure}

The policy network was tested with a simulated user~\cite{schatzmann2006survey} which provides the interaction at the semantic level.
As shown in Figure \ref{fig:SLvsRL}, the first grid points labelled `SL:0' represent the performance of the SL model under various semantic error rates (SER), averaged over 500 dialogues.

The SL model was then further trained using RL at different SERs. 
As the SL model is already performing well, the exploration parameter $\epsilon$ was set to 0.1. The size of the experience replay pool ${\mathcal{L}}$ was 2,000, and the mini-batch size was set to 32. For each update, natural gradient was calculated by eNAC to update the model weights of size $\sim$2600. The total return given to each dialogue was set to $20\times\mathds{1}(\mathcal{D})-T$, where $T$ is the dialogue turn number and $\mathds{1}(\mathcal{D})$ is the success indicator for dialogue $\mathcal{D}$. Maximum dialogue length was set to 30 turns. Return normalisation was used to stabilise training.

The success rate of the SL model can be seen to increase for all SERs during 6,000 training dialogues, spreading between 1-8\% improvement. Generally speaking, the greatest improvement occurs when the SER is most different to the SL training set, which are the higher SER conditions here. In this case, as the semantic hypotheses were more corrupted, the model learned to double-check more on what the user really wanted.
This indicates the model's ability to refine its own behaviour via RL.

\subsection{Policy Network with Real Users} 
\label{sec:real}

Starting from the same SL policy network as in \S  \ref{sec:simulate}, the model was improved via RL using human subjects recruited via AMT.
The policy network was plugged-in to a modular SDS,
comprising the Microsoft's Bing speech recogniser \footnote{www.microsoft.com/cognitive-services/en-us/speech-api.}, a rule-based semantic decoder, the {\it focus} belief state tracker, and a template-based natural language generator.

To ensure the dialogue quality, only those dialogues whose objective system check matched with the user rating were considered \cite{milica_icassp13}.
Based on this, two parallel policies were trained with 200 dialogues.
To evaluate the resulting policies,
policy learning was disabled and a further 110 dialogues were collected with both the SL only and SL+RL models. The AMT users were asked to rate the dialogue quality by answering the question ``\textit{Do you think this dialogue was successful?}" on a 6-point Likert scale and also providing a binary rating on dialogue success. The average quality rating (scaled from 0 to 5) is shown in Table \ref{table:eval} with one standard error. 
The results indicate that the SL-model could work quite well with humans, but was improved by RL on the 200 training dialogues. This demonstrates that on-line RL is a viable approach to adapt a dialogue system to changing environmental conditions.

\begin{table}[h]
\caption{{User evaluation on the policies. Quality: 6-point Likert scale, Success: binary rating. }} 
\begin{center}
\begin{tabular}{ccc}
    \toprule
	\textbf{policy}	& SL & SL+RL  \\ 
    \midrule
	\textbf{Quality (0-5)}    & 3.97 $\pm$ 0.12    &  4.04 $\pm$ 0.12 \\
	\textbf{Success (\%)}    & 94.5 $\pm$ 2.2    &  98.2 $\pm$ 1.2  \\
	\bottomrule

\end{tabular}
\end{center}
\label{table:eval}
  \vspace{-3mm}
\end{table}

\section{Conclusion} \label{sec:conclude}
This paper has presented a two-step development for the dialogue management in SDS using a unified neural network framework,
where the model can be trained on a fixed dialogue dataset using SL and subsequently improved via RL through simulated or spoken interactions. The experiments demonstrated the efficiency of the proposed model with only a few hundred supervised dialogue examples. The model was further tested in simulated and real user settings. In a mismatched environment, the model was capable of modifying its behaviour via a delayed reward signal and achieved better success rate. 

\section*{Acknowledgments}
Pei-Hao Su is supported by Cambridge Trust and the Ministry of Education, Taiwan.

\bibliography{emnlp2016}
\bibliographystyle{emnlp2016}


\end{document}